\documentclass[11pt,a4paper]{article}
\usepackage[hyperref]{emnlp2020}
\usepackage{array}
\usepackage{times}
\usepackage{latexsym}

\usepackage{graphicx}

\usepackage{microtype}
\usepackage[mathscr]{euscript}
\usepackage{bm}
\usepackage{algorithm}
\usepackage{algpseudocode}
\usepackage{amsmath}
\usepackage{multirow}
\usepackage{multicol}
\aclfinalcopy 

\title {An End-to-End Document-Level Neural Discourse Parser \\ Exploiting Multi-Granularity Representations}
  
\author{Ke Shi$^{\dag}$, Zhengyuan Liu$^{\dag}$, Nancy F. Chen\\
  Institute for Infocomm Research, A*STAR, Singapore\\
  \texttt{\{shi\_ke,liu\_zhengyuan,nfychen\}@i2r.a-star.edu.sg} \\}

\date{}

\begin{document}
\maketitle
\newcommand\blfootnote[1]{%
\begingroup
\renewcommand\thefootnote{}\footnote{#1}%
\addtocounter{footnote}{-1}%
\endgroup
}
\begin{abstract}
Document-level discourse parsing, in accordance with the Rhetorical Structure Theory (RST), remains notoriously challenging. Challenges include the deep structure of document-level discourse trees, the requirement of subtle semantic judgments, and the lack of large-scale training corpora. To address such challenges, we propose to exploit robust representations derived from multiple levels of granularity across syntax and semantics, and in turn incorporate such representations in an end-to-end encoder-decoder neural architecture for more resourceful discourse processing. In particular, we first use a pre-trained contextual language model that embodies high-order and long-range dependency to enable finer-grain semantic, syntactic, and organizational representations. We further encode such representations with boundary and hierarchical information to obtain more refined modeling for document-level discourse processing. Experimental results show that our parser achieves the state-of-the-art performance, approaching human-level performance on the benchmarked RST dataset.
\end{abstract}

\section{Introduction}
\begin{figure}
    \begin{center}
      \includegraphics[width=0.44\textwidth]{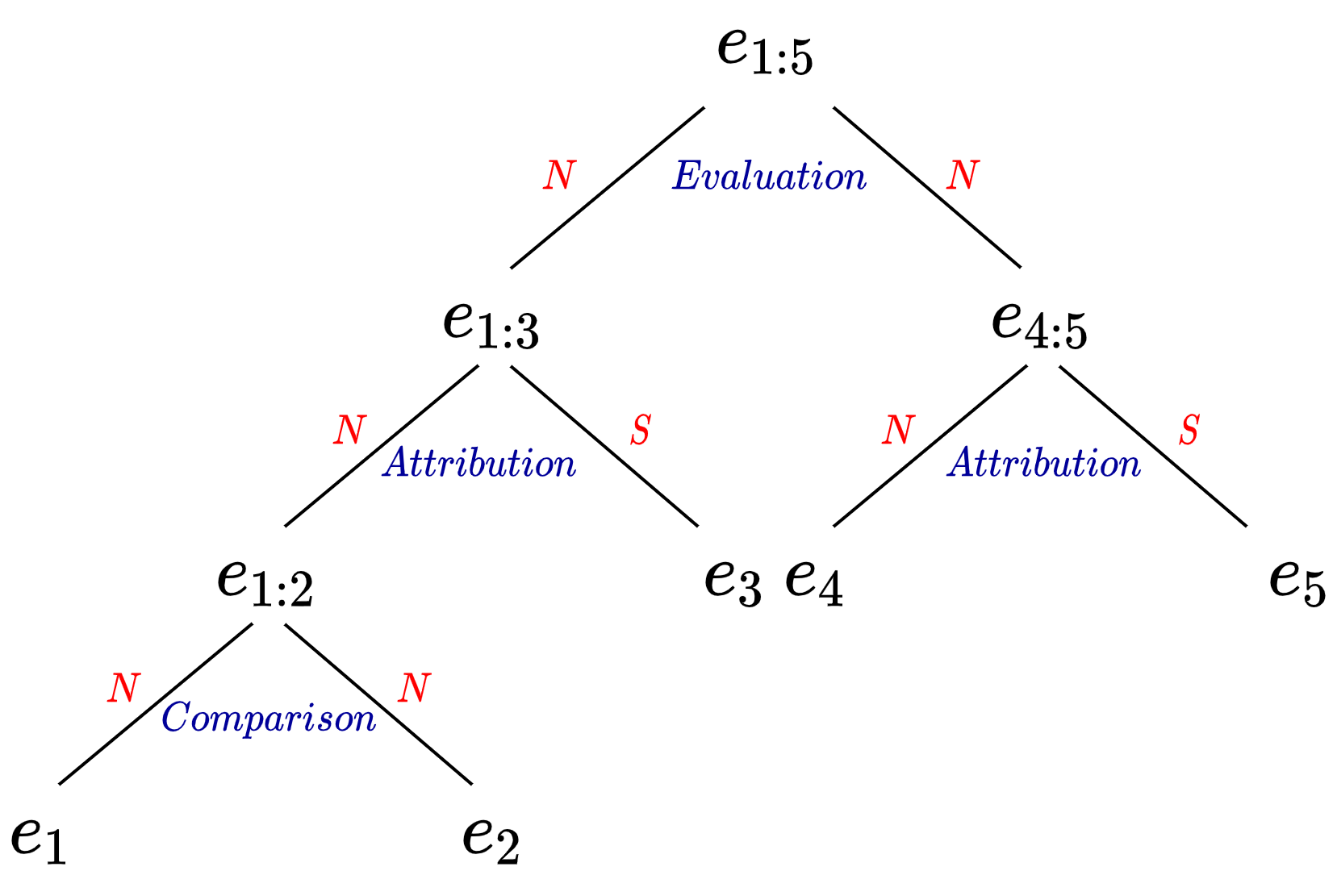}
    \end{center}
    \small{
    $e_1$$[$ The European Community's consumer price index rose a provisional 0.6\% in September from August $]$
    $e_2$$[$ and was up 5.3\% from September 1988, $]$
    $e_3$$[$ according to Eurostat, the EC's statistical agency. $]$
    $e_4$$[$ The month-to-month rise in the index was the largest since April, $]$
    $e_5$$[$ Eurostat said. $]$ 
    }
    \caption{An example of RST discourse tree. $e_i$, $e_{j:k}$, $N$ and $S$ denote elementary discourse units, spans, nucleus and satellite respectively.}
    \label{fig:RST_TREE}
\end{figure}

As a fundamental task in natural language processing (NLP), coherence analysis can benefit various downstream tasks, such as sentiment analysis \citep{choi-etal-2016-document} and document summarization \citep{xu-etal-2020-discourse}. Rhetorical Structure Theory (RST) \citep{mann1988rhetorical} is one of the most influential theories of text coherence, under which a document is represented by a hierarchical discourse tree, which consists of a set of semantic units organized in the form of a dependency structure, labeled with their rhetorical relations.\blfootnote{$^{\dag}$Equal contribution.}
As shown in Figure \ref{fig:RST_TREE}, the leaf nodes of an RST discourse tree are basic text spans called Elementary Discourse Units (EDUs), and the EDUs are iteratively connected by rhetorical relations (\emph{e.g.,} \textit{Elaboration} and \textit{Contrast}) to form larger text spans until the entire document is included. 
The rhetorical relations are further categorized to \textit{Nucleus} and \textit{Satellite} based on their relative importance, in which \textit{Nucleus} corresponds to the core part(s) while \textit{Satellite} corresponds to the subordinate part. While manual coherence analysis under the RST theory is labor-intensive and requires specialized linguistic knowledge, a discourse parser serves to automatically transform a document into a discourse tree. Document-level discourse parsing consists of three sub-tasks: hierarchical span splitting, rhetorical nuclearity determination, and rhetorical relation classification.

Models for RST-style discourse parsing have made much progress in the past decade. While statistical methods utilize hand-crafted lexical and syntactic features \citep{soricut2003sentence,sagae2009analysis,joty2013combining,feng2014linear,heilman2015fast}, data-driven neural approaches reduce feature-engineering labor by effective representation learning, and are capable of characterizing implicit semantic information. Neural networks are first used as feature extractors along with traditional shift-reduce approaches \citep{ji2014representation} or dynamic programming approaches \citep{li2016discourse}. Then, \citet{yu2018transition} bridges the gap between neural and traditional methods by an end-to-end transition-based neural parser via an encoder-decoder architecture. Recently, pointer networks are introduced to achieve linear-time complexity, and models with top-down parsing procedures achieve favorable results on sentence-level discourse analysis tasks \citep{lin-etal-2019-unified, liu2019hierarchical}.

However, there is still much space for improvement in document-level discourse parsing.
First, compared to sentence-level parsing, document-level parsing is more challenging due to the deeper tree structures and longer dependencies among EDUs: in the benchmark dataset RST Discourse Tree Bank (RST-DT) \cite{carlson2002rst}, the average EDU number at the document level is 56, which is 20 times larger than that of sentence-level parsing.
Thus modeling context information across a long span is essential, especially if considering a top-down parsing procedure where poor accuracy at the top of the tree will propagate toward the leaf nodes.
Second, the three sub-tasks of discourse parsing strongly rely on nuanced semantic judgments, which require comprehensive contextual representation with various types of linguistic information. Take discourse relation classification for example, explicit relations are overtly signaled by a connective word such as ``although'' and ``because'', which can be determined by lexical and syntactic features. However, this approach can not be readily adapted to implicit discourse relations determination, as it requires high-order features with semantic information.
Moreover, to compensate for the lack of large-scale corpora, prior work in neural modeling has leveraged inductive biases through syntactic features such as part-of-speech tagging to improve performance. However, such models still suffer from insufficient linguistics information from the lack of data, thus they are incapable of acquiring deeper and richer contextual representations useful for discourse processing.

In this paper, to tackle the aforementioned challenges, we propose a document-level neural discourse parser with robust representation modeling at both the EDU and document level, based on a top-down parsing procedure. To take advantage of widely-adopted vector representations that encode rich semantic information, we first exploit a large-scale pre-trained language model as a contextual representation backbone.  Then we incorporate boundary information with implicit semantic and syntactic features to the EDU representations, and introduce a hierarchical encoding architecture to more comprehensively characterize global information for long dependency modeling.
To improve inference accuracy and alleviate the aforesaid error propagation problem, we present breadth-first span splitting to propose a layer-wise beam search algorithm.

We train and evaluate our proposed model on the benchmark corpus RST-DT\footnote{https://catalog.ldc.upenn.edu/LDC2002T07} \citep{carlson2002rst}, and achieve the state-of-the-art performance on all fronts, significantly surpassing previous models while approaching the upper bound of human performance. We also conduct extensive experiments to analyze the effectiveness of our proposed method.

\section{Related Work}
RST discourse parsing has been in the spotlight since \citep{marcu1997rhetorical,corston1998beyond}. Statistical models are dominant in initial studies. \citet{soricut2003sentence} proposed a learning-based probabilistic model to build the discourse trees, then the shift-reduce based framework was introduced to discourse parsing by \citet{sagae2009analysis}. Later greedy transition-based shift-reduce parsers achieved cutting-edge performances \citep{li2014text,ji2014representation,heilman2015fast,wang2017two}. 
Different from shift-reduce parsers, \citep{joty2013combining,feng2014linear,li-etal-2014-recursive} employed conditional random field approaches to seek globally optimal results, where two EDUs with the highest relational probability were merged into one span iteratively to generate the discourse tree.
In addition, linguistic features have been demonstrated to be effective in RST discourse parsing. Classic features adopted by previous work \citep{sagae2009analysis,hernault2010hilda,joty2013combining,feng2014linear,heilman2015fast,wang2017two} include: (1) $N$-gram features, which is a lexical dictionary, describe the discourse cues (\emph{e.g.,} ``because'', ``however''); (2) Syntactic features such as part-of-speech (POS) tags; (3) Organizational features, which illustrate the textual organization including the number of EDUs and length of tokens, as well as distances of the units from the beginning and the end of text spans; (4) Dominance sets and lexical chains which show the dominance relation and indicate topic shifts respectively. Furthermore, \citet{li2014text} explored the benefits of dependency structures in discourse parsing. 

Recently, neural networks have been making inroads into discourse analysis frameworks. \citet{li2016discourse} proposed an attention-based hierarchical neural network model, which employed a Bi-LSTM network for obtaining compositional semantic representations. 
\citet{yu2018transition} improved the performance by integrating neural syntax features into the greedy transition-based parser. In addition, \citet{lin-etal-2019-unified} and their follow-up work \citet{liu2019hierarchical} successfully explored encoder-decoder neural architectures on sentence-level discourse analysis, with a top-down parsing procedure. \citet{zhang-etal-2020-top} evaluated the effectiveness of top-down parsing compared with a bottom-up approach. More recently, \citet{kobayashi-etal-2020-Top} proposed a multi-stage parsing process from document-level and paragraph-level to sentence-level. \citet{liu-etal-2020-multilingual-neural} investigated cross-lingual representations and EDU-level translation on multilingual RST discourse parsing.

In this work, we show how incorporating different levels of granularity of implicit linguistic features and hierarchically modeling the content from the EDU to the document level, can improve the performance on the more challenging document-level RST discourse parsing task.

\section{The Proposed Model}

\begin{figure*}
    \centering
    \includegraphics[width=15cm]{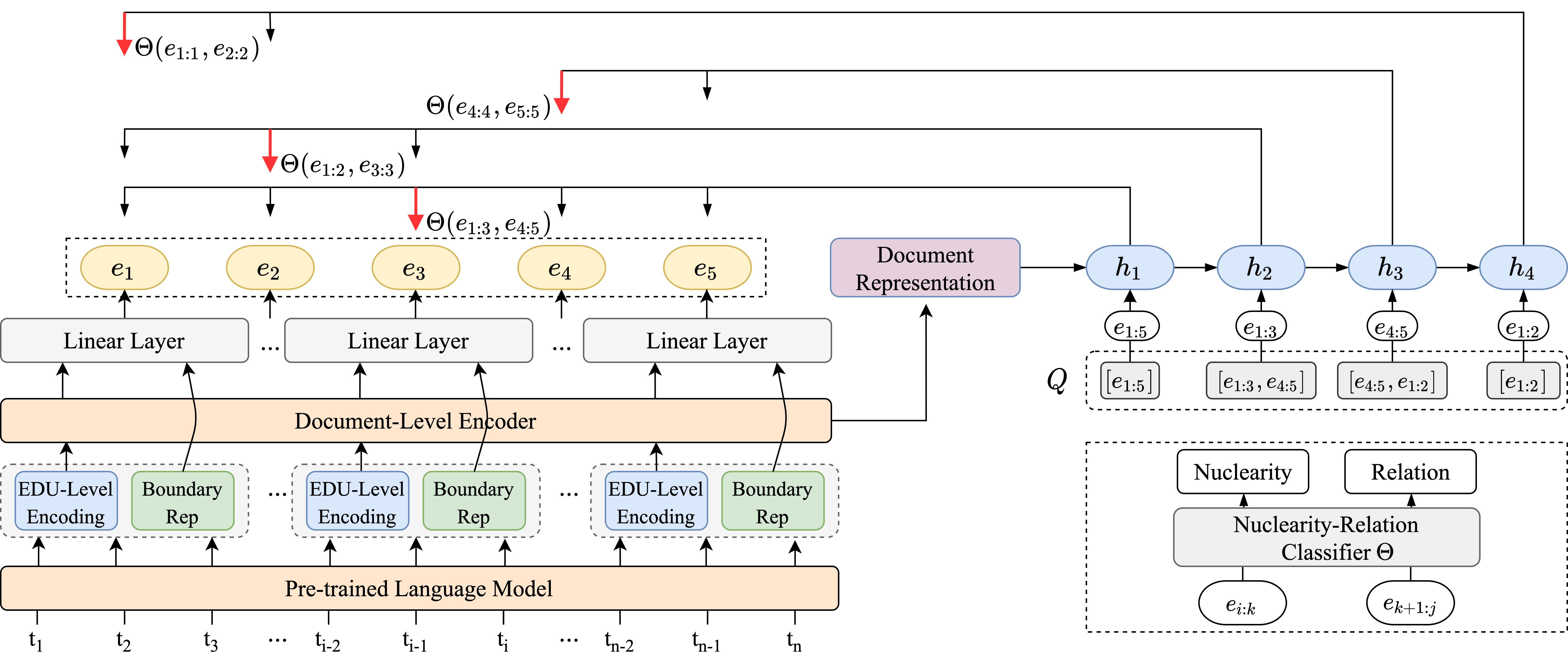}
    \caption{Overview of the proposed neural architecture for document-level discourse parsing. The left part is the encoder and the right part is the decoder. $t$, $e$ and $h$ denote input token, encoding hidden state, decoding hidden state respectively. $\Theta$ refers to nuclearity-relation classification. Queue $Q$ is initialized with the span $e_{1:m}$, and maintained by the decoder to track the top-down span splitting. With each splitting pointer $k$, sub-spans $e_{i:k}$ and $e_{k+1:j}$ are fed to a classifier for nuclearity and relation determination.}
    \label{fig:Architecture}
\end{figure*}

Given a document containing EDU segments as input, the discourse parser iteratively splits it to spans with EDUs, and constructs a hierarchical tree in a transition-based top-down manner, then it determines the nuclearities and relations between these spans.
Our framework consists of three components: (\romannumeral1) Hierarchical Encoder; (\romannumeral2) Attention-based Span Splitting Decoder; (\romannumeral3) Nuclearity-Relation Classifier. The overview of the proposed architecture is shown in Figure \ref{fig:Architecture}.

\subsection{Hierarchical Encoder}
\label{ssec:encoder}
\noindent\textbf{EDU-Level Encoding}:
Given a document containing $n$ tokens $T=\{t_1,...,t_n\}$, the EDU-level encoder converts them to distributional semantic vectors as EDU-level representations. Here, we select XLNet \citep{yang2019xlnet} as the backbone. Since XLNet has strong high-order and long dependency modeling capability, and the supported input length is up to 2048, it is more suitable than other widely used large-scale pre-trained language models like ELMo \citep{peters-etal-2018-deep} or BERT \citep{Devlin_2019} to characterize the deep structures of document-level discourse trees. 
With the token embeddings $\widetilde{T}=\{\widetilde{t}_1,...,\widetilde{t}_n\}$ produced by the language model, EDU-level representations $C = \{c_1,...,c_m\}$ are obtained by averaging the tokens embeddings in respective EDUs, where $m$ is the EDU number in the document. By averaging at the token level, we expect that each $c_i$ can aggregate and represent the semantic information of an EDU span. Besides, weighted schemes are shown to be effective for downstream tasks \cite{Zhang*2020BERTScore:}. Therefore, we also explore self-attentive and GRU-attentive methods for the weighted EDU-level aggregation (see Section \ref{ssec:weighted_representation} for comparisons).

\noindent\textbf{Document-Level Encoding}: To capture the dependencies among the EDUs and obtain the document-level representation for global information, the encoded EDUs $C$ are fed into a multi-layer Bi-GRU component for sequential modeling. The context-aware EDU-level representations $V=\{v_1,..., v_m\}$ are obtained by concatenating the forward and the backward hidden states $v_i=[\stackrel{\rightarrow}{v_i}; \stackrel{\leftarrow}{v_i}]$. Meanwhile, the document representation can be obtained from the last hidden state of Bi-GRU and it is fed to the decoder in Section \ref{ssec:decoder} as the initial hidden state to provide a holistic view of the document.\footnote{We also adopt Transformer \citep{vaswani2017attention} as an alternative component for document-level encoding, and the document representation is obtained by averaging the last-layer hidden states (see Section \ref{ssec:weighted_representation} for comparisons).}

\noindent\textbf{Incorporating Boundary Information}: In non-neural discourse parsing methods, shallow but effective lexical features are widely used. For instance, discourse relations are often triggered by specific words or phrases such as subordinating conjunctions (\emph{e.g.,} ``because'', ``although''), coordinating conjunctions (\emph{e.g.,} ``or'', ``but''), and discourse adverbials (\emph{e.g.,} ``however'', ``for example''). Meanwhile, syntactic information like POS tags and certain structure information (\emph{e.g.,} spans ended with punctuation marks) are also useful in constructing the tree.  Intuitively, these features are position-related and distributed in the boundaries of EDUs. Therefore, we introduce the boundary representations $G=\{g_1,...,g_m\}$, which convey such linguistic features implicitly. Each $g_i$ is composed of token embedding vectors in $\widetilde{T}$ at both ends of EDU $i$. Then, we obtain the enhanced EDU representations $E=\{e_1,...,e_m\}$ by compressing the contextual-aware $V$ and boundary information $G$ with a shared linear layer.

\begin{equation}
\label{boundary_equation}
    e_i = W_e([v_i;g_i]) + b_e
\end{equation}
where $;$ denotes the concatenation operation. $W_e$ and $b_e$ are the trainable parameter matrix and bias.

\subsection{Attention-Based Span Splitting Decoder}
\label{ssec:decoder}
The decoding phase is conducted in a transition-based procedure to split spans of EDUs to form the tree structure. 
The depth-first manner is widely used for top-down approaches in parsing tasks \cite{lin-etal-2019-unified}. However, if parsing accuracy is poor at higher levels, errors could propagate to lower levels. To alleviate this problem, we present a layer-wise beam search algorithm (see Section \ref{sec:layer_beam_search}), and conduct the decoding procedure in a \textbf{top-down breadth-first manner}. Figure \ref{fig:Architecture} illustrates the parsing steps of the example in Figure \ref{fig:RST_TREE}: the decoder maintains a queue $Q$, which is initialized by the span that contains all EDUs $e_{1:m}$. At each decoding step $t$, the span $e_{i:j}$ at the head of $Q$ is parsed into two sub-spans $e_{i:k}$ and $e_{k+1:j}$ where $i\leq k < j$ based on the predicted splitting position. Afterwards, spans that need further parsing (spans containing more than one EDU) will be appended into the tail of $Q$ to maintain the Breadth-First process, then the decoder iteratively parses the spans until $Q$ is empty to form a discourse tree. Note that this breadth-first procedure is a general design which can be applied in other tree structure tasks such as dependency parsing. 

The pointer network \citep{vinyals2015pointer} is used to predict the splitting position, according to the computed attention scores on encoded EDU representations $e_i$. In the decoder component, we employ a unidirectional GRU layer. The last hidden state of the document-level encoder is used to initialize the hidden states $h_0$ of the GRU. At each decoding step, the input span representation is calculated from taking the average of the respective EDU representations (\emph{i.e.} $mean(e_i,...,e_k)$ for $e_{i:k}$ where $0<i\leq k\leq m$). 
Next, hidden state $h_t$ is produced by the GRU with the previous hidden state $h_{t-1}$ and span representation $e_{i:k}$. Then, the splitting prediction is based on the attention mechanism \citep{bahdanau2014neural}, and the attention scores are computed over the corresponding EDU representations (\emph{i.e.} $e_i,...,e_j$ for $e_{i:j}$), which is a softmax distribution over the input span.
\begin{equation}
    s_{t,u} = \sigma (h_t, e_u) \ \  \mathbf{for}\ \  u = i...j
\end{equation}
\vspace{-0.8cm}
\begin{equation}
    a_t = softmax(s_t) = \frac{exp(s_{t,u})}{\sum_{u=i}^jexp(s_{t,u})}
\end{equation}
\noindent where $\sigma(x,y)$ is the dot product used as attention scoring function.

\subsection{Nuclearity-Relation Classifier}
In each decoding step, after the span $e_{i:j}$ is split into two sub-spans $e_{i:k}$ and $e_{k+1:j}$, a bi-affine classifier is adopted to predict their nuclearity and relation labels.
Following \citet{lin-etal-2019-unified}, we attach the nuclearity labels \textit{Nucleus-Satellite (NS)}, \textit{Satellite-Nucleus (SN)} and \textit{Nucleus-Nucleus (NN)} to the relation labels.

The classifier contains two neural layers, the first is a dense layer with Exponential Linear Unit (ELU) activation, which projects the span representation $e_l$ and $e_r$ to latent features $\widetilde{e}_l$ and $\widetilde{e}_r$ with the dimensions $d$, where $e_l$ and $e_r$ are the means of corresponding EDU representations in the left span $e_{i:k}$ and right span $e_{k+1:j}$:
\begin{equation}
    \widetilde{e}_l=ELU(e_l^TU_1);\  \widetilde{e}_r = ELU(e_r^TU_2)
\end{equation}
Following \citet{DBLP:conf/iclr/DozatM17}, these two latent features are fed into a bi-affine layer with softmax activation:
\begin{equation}
\begin{aligned}
    P_\theta(y|X) = softmax(\widetilde{e}_l^TW_l + \widetilde{e}_l^TW_{lr}\widetilde{e}_r \\ + \widetilde{e}_r^TW_r + b)
\end{aligned}
\end{equation}
\noindent where $W_l\in\mathcal{R}^{d\times R}$; $W_r\in\mathcal{R}^{d\times R}$ and $W_{lr}\in\mathcal{R}^{d\times d\times R}$ are the weights and bias $b \in \mathcal{R}^R$.

\subsection{Training Loss}
The objective of our parser is to minimize the total loss of parsing the correct tree structure and predict the corresponding nuclearity and relation labels. The structure loss $\mathcal{L}_s$ is the cross entropy of the span splitting, and the loss of label prediction $\mathcal{L}_l$ is the cross entropy loss for the nuclearity-relation classifier:
\begin{equation}
    \mathcal{L}_s(\theta_s) = - \sum_{t=1}^TlogP_{\theta_s}(y_t|y_1,...,y_{t-1},X)
\end{equation}
\vspace{-0.4cm}
\begin{equation}
    \mathcal{L}_l(\theta_l) = - \sum_{m=1}^M\sum_{r=1}^RlogP_{\theta_l}(y_m=r|X)
\end{equation}
\noindent where $\theta_s$ and $\theta_l$ are the parameters of the pointer network and classifier respectively, $T$ is the total number of spans, and $y_1,...,y_{t-1}$ denote the subtrees that have been generated in the previous steps. $M$ is the number of spans with at least two EDUs, and $R$ is the total number of nuclearity-relation labels. 

The total loss with $L_2$-regularization is:
\begin{equation}
     \mathcal{L}_{total}(\theta^{*}) = \mathcal{L}_s(\theta_s) + \mathcal{L}_l(\theta_l) + \lambda||\theta^{*}||_2^2
\end{equation}
\noindent where $\lambda$ is the regularization strength and $\theta^{*}$ refers to all learning parameters of the model.

\section{Layer-Wise Beam Search}
\label{sec:layer_beam_search}
In the top-down decoding process described in Section \ref{ssec:decoder}, a regular decoder uses the greedy algorithm to choose a point to split one span into two sub-spans, where the decision is made by only looking at the highest probability at each decoding step. However, this approach could lead to locally optimal decisions when the splitting in the upper layer generates two inferior sub-spans in the subsequent decoding step, which is what we have observed in preliminary experiments and analyses.

The naive beam search algorithm used in sequence-to-sequence models selects multiple alternatives at each decoding step based on previous conditional probability, which can enlarge the search space compared with the greedy algorithm.
However, it can not be adopted in our model for tree structure inference as the decoding probabilities of different spans in the same layer are independent.
For instance, in Figure \ref{fig:RST_TREE}, after span $e_{1:5}$ is spilt into span $e_{1:3}$ and span $e_{4:5}$, decoding span $e_{1:3}$ and decoding span $e_{4:5}$ are independent. Therefore, we assume the probabilities of decoding all spans in one layer only condition on their corresponding parent layers, and propose a beam search algorithm under the layer-wise conditional probability:
\begin{equation}
\begin{aligned}
    P_{y_{layer}^1,...,y_{layer}^t}= \sum_{i=1}^tP(y_{layer}^i|y_{layer}^1,\\ ...,y_{layer}^{i-1},X)
\end{aligned}
\end{equation}
\vspace{-0.6cm}
\begin{equation}
        \label{equ:Player}
      P_{y_{layer}^i} = \sum_{j=1}^klog{P_s}_i^j + \sum_{j=1}^klog{P_r}_i^j
\end{equation}
where $P_{y_{layer}^i}$ refers to the log probabilities of decoding all spans in the $i^{th}$ layer. ${P_s}_i^j$ and ${P_r}_i^j$ are the splitting and relation probabilities of decoding the $j^{th}$ spans in layer $i$. $t$ is the number of layers, and $k$ is the total number of spans in the $i^{th}$ layer.

Algorithm \ref{algorithm} shows the details of the proposed layer-wise beam search algorithm. During the inference process, the decoder maintains a queue $Q_{current\underline{\ }layer}$ which contains $K$ span set candidates with the highest probabilities for the current layer, where $K$ is the beam size and the `span set' refers to all spans in one layer (\emph{e.g.,} considering the example in Figure \ref{fig:RST_TREE}, with a beam size of 2, the $Q_{current\underline{\ }layer}$ may contain two span set candidates such as $[[e_{1:3},e_{4:5}];[e_{1:2},e_{3:5}]]$ in the second layer). 
For each span set candidate in $Q_{current\underline{\ }layer}$, the decoder first parses it into sub-span sets for the next layer, then stores top $K$ possibilities in $Q_{next\underline{\ }layer}$, where the probabilities are calculated as in Equation \ref{equ:Player}. After parsing all span set candidates in $Q_{current\underline{\ }layer}$, the decoder selects top $K$ candidates from $K^2$ combinations in $Q_{next\underline{\ }layer}$, and $Q_{current\underline{\ }layer}$ is updated by the selected $K$ span set candidates. Then the decoder iteratively parses candidates in $Q_{current\underline{\ }layer}$ until all spans only contain one EDU.

\begin{algorithm}[t]
\linespread{1.1}
\small
    Beam Size = $K$; $P_s$ and $P_r$ are the probabilities of span splitting and relation classification.
    \caption{Layer-Wise Beam Search}
    \begin{algorithmic}[1]
        \State Initialize $Q_{current\underline{\ }layer} = [[e_{1:m}]]$
        \State KeepSplit = True
        \While {KeepSplit}
            \State $Q_{next\underline{\ }layer}$ $\leftarrow$ Empty List
            \For {span set candidate $S_{cand}$ in $Q_{current\underline{\ }layer}$}
                \State $Q_{tmp}$ $\leftarrow$ Empty List
                \For {span in $S_{cand}$}
                    \If {span only contains 1 EDU}
                        \State Continue
                    \EndIf
                    \State Split span with $P_s$
                    \State Predict $P_r$ of sub-span pairs
                    \State Select top-$K$ sub-span pairs by $logP_s+logP_r$
                    \State $Q_{tmp} = $ Combine candidate sub-span pairs with items in $Q_{tmp}$
                \EndFor
                \State Add top $K$ sub-span sets in $Q_{tmp}$ to $Q_{next\underline{\ }layer}$ 
            \EndFor
            \State $Q_{current\underline{\ }layer}$ $\leftarrow$ top $K$ span set candidates for the next layer from $Q_{next\underline{\ }layer}$, where probabilities are normalized by the number of generated spans.
            \If {$Q_{current\underline{\ }layer}$ is empty}
                \State  KeepSplit = False
            \EndIf
        \EndWhile
    \end{algorithmic}
    \label{algorithm}
\end{algorithm}

\section{Experiments}

\subsection{Dataset}
We followed previous studies \cite{ji2014representation,li2014text,feng2014linear,li2016discourse,wang2017two, braud-etal-2017-cross-lingual,yu2018transition}, trained and evaluated our model on the standard RST Discourse Treebank (RST-DT) corpus \citep{carlson2002rst}. RST-DT contains 385 documents collected from the Wall Street Journal (WSJ), in which 347 documents for training and 38 for testing. We randomly sampled 35 documents from the training set as a validation set. For the parsing task, we conducted our experiments on documents with gold EDU segmentation.

\subsection{Evaluation Metrics}
We applied the standard micro-averaged F1 scores on \textit{Span} (\textbf{S}), \textit{Nuclearity-Satellite} (\textbf{NS}), \textit{Relation} (\textbf{R}) and \textit{Full} under RST Parseval evaluation procedure proposed by \citet{marcu2000rhetorical}, where \textit{Span} describes the capability of constructing the tree structure, \textit{Nuclearity-Satellite} and \textit{Relation} assess the ability to indicate the nuclearity and judge the discourse relations respectively. We also adopted \textit{Full} to evaluate the tree structure together with both \textit{Nuclearity-Satellite} and \textit{Relation} as in \cite{morey-etal-2017-much}. Following previous studies, we adopted the same 18 relations defined in \cite{carlson2001discourse}. In addition, we also evaluated our proposed models by the Labelled Attachment Decisions (Original Parseval) \citep{morey-etal-2017-much} for better comparison with previous studies.

\subsection{Training Configuration}
The proposed model was implemented in PyTorch \cite{paszke2019pytorch}. We used \textit{`xlnet-base-cased'} in \citet{Wolf2019HuggingFacesTS} and fine-tuned the last 4 layers during training. Documents were tokenized with the Byte Pair Encoding scheme and
fine-tuned on the validation set (see Table \ref{parameter-table} for details). We trained the model for 30 epochs (7.5 hours), and selected the best checkpoint on the validation set for evaluation. For each round of evaluation, we repeated the training 6 times with different random seeds and averaged the scores. The trainable parameter size was 67M, where 31M parameters were from fine-tuning XLNet. All experiments were run on 
a Tesla V100 GPU with 16GB memory.

\begin{table}
\linespread{1.1}
\centering
\small
\begin{tabular}{rc}
\hline
\textbf{Parameter} & \textbf{Value} \\
\hline
Token Embedding Dimension & 768 \\
Document-Level Bi-GRU Dimension & 384 \\
Decoder Uni-GRU Dimension & 768 \\
\hline
Dropout Rate & 0.5 \\ 
Optimization Algorithm & Adam \\
Batch Size & 3 \\ 
Learning Rate & 0.001 \\
Weight Decay & 0.0005 \\
Layer-Wise Beam Search Size & 5 \\
\hline
\end{tabular}
\caption{Training configuration}
\label{parameter-table}
\end{table}

\begin{table*}[ht]
\linespread{1.2}
\footnotesize
\begin{center}
\begin{tabular}{p{4.0cm}|p{0.75cm}<{\centering}p{0.75cm}<{\centering}p{0.75cm}<{\centering}p{0.75cm}<{\centering}|p{0.75cm}<{\centering}p{0.75cm}<{\centering}p{0.75cm}<{\centering}p{0.75cm}<{\centering}}
\hline
\multirow{2}*{\textbf{Model}} & \multicolumn{4}{c|}{\textbf{RST Parseval}} & \multicolumn{4}{c}{\textbf{Original Parseval}} \\ 
\cline{2-9}
~ & \textbf{S} &\textbf{NS} & \textbf{R}& \textbf{Full}& \textbf{S} &\textbf{NS} & \textbf{R}& \textbf{Full}\\
\hline
Traditional Models:&&&&&&&&\\
\citep{ji2014representation}* & 82.0 &68.2 & 57.8 & 57.6 & 64.1 & 54.2 & 46.8 & 46.3\\
\citep{feng2014linear}* & 84.3 &69.4 & 56.9 & 56.2 & 68.6 & 55.9 & 45.8 & 44.6\\
\citep{surdeanu2015two}* & 82.6&67.1 &55.4 &54.9 & 65.3 & 54.2 & 45.1 & 44.2\\
\citep{joty2015codra}* &82.6& 68.3& 55.8 &55.4 & 65.1 & 55.5 & 45.1 & 44.3\\
\citep{hayashi2016empirical}* &82.6& 66.6& 54.6&54.3 & 65.1 & 54.6 & 44.7 & 44.1 \\
\hline
Neural Models:&&&&&&&\\
\citep{braud2016multi-view}* &79.7& 63.6& 47.7 &47.5& 59.5 & 47.2 & 34.7 & 34.3\\
\citep{li2016discourse}* & 82.2 & 66.5 & 51.4 &50.6 & 64.5 & 54.0 & 38.1 & 36.6\\
\citep{braud-etal-2017-cross-lingual}* &  81.3 & 68.1 & 56.3&56.0 &  61.9 & 53.4 & 44.5 & 44.0\\
\citep{yu2018transition} & 85.5 & 73.1 & 60.2&59.9 & - & - & - & -\\
\citep{lin-etal-2019-unified} & 85.0 & 70.4 & 57.5&57.1 & - & - & - & -\\
\cite{zhang-etal-2020-top} & - & - & - & - &  67.2 & 55.5 & 45.3 & 44.3\\
\citep{kobayashi-etal-2020-Top} & 87.0 & 74.6& 60.0& - & - & - & - & -\\
\hline
Our Model & \textbf{87.9}  & \textbf{75.4} & \textbf{63.3} &\textbf{62.9}  & \textbf{76.2}  & \textbf{64.8} & \textbf{53.5} &\textbf{52.5}\\ 
+ Layer-Wise Beam Search & \textbf{88.2}  & \textbf{75.7} & \textbf{63.7} & \textbf{63.3} & \textbf{76.3} & \textbf{65.0} & \textbf{53.9} & \textbf{52.9}\\
\hline
Human* & 88.3  & 77.3 & 65.4 & 64.7 & 78.7  & 66.8 & 57.1 & 55.0\\
\hline
\end{tabular}
\end{center}
\caption{Micro-F1 results on the RST test set of baseline models and our proposed model, * denotes the results are from \citet{morey-etal-2017-much}. RST Parseval \cite{marcu2000rhetorical} and Original Parseval \cite{morey-etal-2017-much} are used.}
\label{Micro-F1 results}
\end{table*}

\subsection{Test Results}
Table \ref{Micro-F1 results} shows the results of our proposed model and a series of baselines on the RST test set. \citet{morey-etal-2017-much} calculated the human agreement and replicated 8 successful studies to uniformly evaluated them by micro-averaged F1 score under both RST Parseval and Original Parseval, so we list them in Table \ref{Micro-F1 results} for ease of comparison.
In addition, we adapted the current state-of-the-art model on sentence-level discourse parsing \citep{lin-etal-2019-unified} for the document-level task as another competitive baseline.

As shown in Table \ref{Micro-F1 results}, neural models have gained traction in recent years, especially through end-to-end transition-based learning \citep{yu2018transition} and multi-stage parsing process \citep{kobayashi-etal-2020-Top}. However, the performance of nuclearity and relation classification is still far from human performance. Our model surpasses prior work with a considerable margin under RST Parseval on all sub-tasks\footnote{Micro-F1 score of the reported model on the validation set is 87.7 (span), 75.9 (nuclearity-satellite), 64.0 (relation) and 63.5 (full).}, especially on \textit{Relation} and \textit{Full}, and approaches human performance. Minor yet consistent improvements could be made by using the layer-wise beam search for further optimization\footnote{In our setting, applying layer-wise beam search only increase 1.25\% space in the inference stage (8223 MB vs 8343 MB), and the trainable parameters are the same. The inference time complexity is $O(nk)$. When beam size $k$ is set of 5, it leads to performance improvement (0.3 F1 on all aspects) of our proposed model.}. 

Evaluations under Original Parseval show more significant improvement, compared with the latest work of \citet{zhang-etal-2020-top}, our base model obtained 9.0, 9.3, 5.2, and 8.2 F1 score increment on the 4 aspects respectively\footnote{\citet{zhang-etal-2020-top} utilized GloVe embeddings in the proposed model, while they reported that ELMo could only improve the performance about 0.6 F1 score.}, and the proposed layer-wise beam search further improved the result to approach the human-level performance.

\section{Analysis}
A series of experiments under RST Parseval were conducted to verify the rationale of our architecture design for each component and the effectiveness of our proposed model.

\subsection{Effect of Boundary Information}
\label{ssec:Boundary Information}
We fed our model with only one-side of the boundary (left or right) information to evaluate their respective contributions. In particular, we assume the left boundaries epitomize more syntactic and semantic information such as conjunctions, while the right boundaries embody more structure information such as punctuation marks. As shown in Table \ref{left-right-analysis}, both types of boundary information achieve improvements: right boundary information contributes to higher tree structure prediction accuracy, while left boundary information benefits relation prediction.

We also evaluated the effect of POS information, which is widely used in previous studies. We extracted the POS tags of the boundary tokens by using Stanford CoreNLP \footnote{https://stanfordnlp.github.io/CoreNLP/} toolkit v4.0 and replaced boundary information with POS embeddings of dimension size 30. However, POS tagging information results in trivial improvement compared to the base model (see Table \ref{left-right-analysis}). Such findings suggest that compared to POS tags at boundary positions, vector representations of boundary information embodies richer information suitable for document-level discourse processing.

\begin{table}[t!]
\linespread{1.1}
\centering
\small
\begin{tabular}{lcccc}
\hline
\textbf{Model} & \textbf{S} &\textbf{NS} & \textbf{R}& \textbf{Full} \\
\hline
Only Add Left Boundary \\ Information & 86.8 & 73.7 & 61.1 & 60.9\\
Only Add Right Boundary \\ Information & 87.3 & 73.4 & 60.5 & 60.1\\
\hline
\hline
Only Add Boundary \\ POS-Tagging Information & 86.2 & 72.6 & 59.7 & 59.4\\
\hline
w/o Boundary Information & 86.0 & 72.1 & 59.6 & 59.2\\
\hline
\end{tabular}
\caption{Micro-F1 scores on different configurations of utilizing boundary information.}
\label{left-right-analysis}
\end{table}

\subsection{Visualization Analysis of Boundary Information}
To further study what boundary information is learned by the model, we analyzed their latent features via decomposition and visualization. The vectors of boundary tokens of EDUs (\emph{i.e.} $g_i$ in Equation \ref{boundary_equation}) were extracted from all the test samples, projected to 2 dimensions for better visualization via principal component analysis (PCA), and finally colored with their corresponding linguistic or structural characteristics. 

\begin{figure}[ht]
    \begin{center}
      \includegraphics[width=5.6cm]{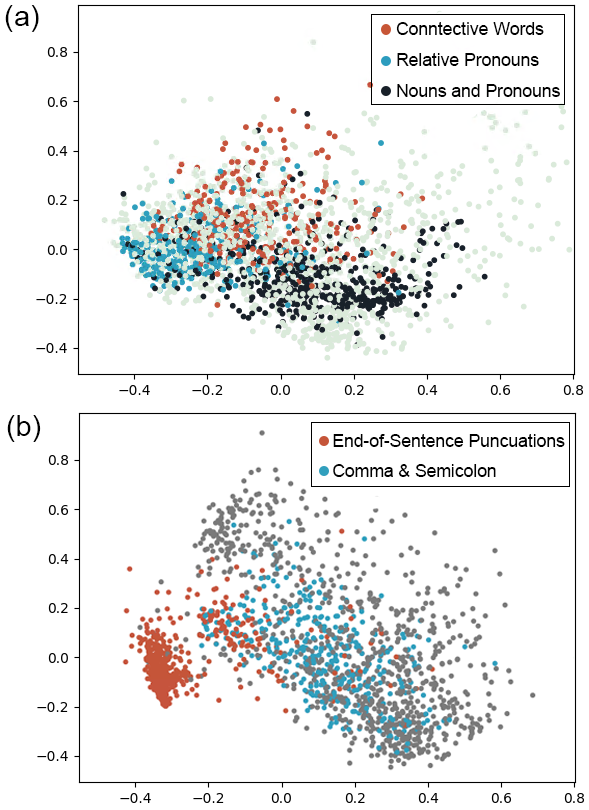}  
    \end{center}
    \caption{Visualization of (a) left boundary vectors and (b) right boundary vectors through PCA.}
    \label{fig:boundary_vis}
\end{figure}

Interestingly, we found that points of the same color showed a tendency of clustering rather than displaying a random distribution. The color-coded clusters in Figure \ref{fig:boundary_vis}a represent the left boundary information, and roughly correspond to connective words (purple), relative pronouns (cyan), pronouns and nouns (black), suggesting their latent features are associated with syntactic and semantic information.
Points in Figure \ref{fig:boundary_vis}b represent the latent feature of the right boundary, where the red cluster mostly consists of punctuation marks placed at the end of a sentence (\emph{e.g.,} period and question mark). This observation is aligned with our hypothesis in Section \ref{ssec:Boundary Information}.

\subsection{Effect of Fine-Tuning Language Models}
Pre-trained language models have demonstrated the effectiveness of incorporating language priors, and downstream tasks can benefit largely by adopting fine-tuning. Table \ref{pretrain-analysis} shows the comparison between using ELMo and XLNet as the representation backbone.\footnote{In our settings, we did not use BERT in the encoding component due to its 512 length limitation.} While they perform nearly the same without fine-tuning, the Transformer-based model (XLNet) obtains more improvement after fine-tuning than the LSTM-based model (ELMo); the former is also more computationally efficient. This result supports our hypothesis that the self-attentive contextual representations in XLNet are more suitable for characterizing deep structures and long dependencies in document-level discourse parsing.

\begin{table}[t!]
\linespread{1.1}
\centering
\small
\begin{tabular}{p{3.0cm}cccc}
\hline
\textbf{Model} & \textbf{S} &\textbf{NS} & \textbf{R} & \textbf{Full}\\
\hline
ELMo (Fixed) & 85.4 & 70.7 & 57.9 & 57.2\\
XLNet (Fixed) & 85.5 & 71.1 & 58.1 & 57.5\\
ELMo (Fine-Tuned) & 86.3 & 72.3 & 60.9 & 60.3\\
XLNet (Fine-Tuned) & 87.9  & 75.4 & 63.3 & 62.9\\ 
\hline
\end{tabular}
\caption{Micro-F1 scores on different pre-trained language model configurations.}
\label{pretrain-analysis}
\vspace{-0.3cm}
\end{table}

\subsection{Effect of Span Encoding}
\label{ssec:weighted_representation}
\noindent\textbf{EDU-Level Encoding}: To evaluate the encoding methods from the token level to the EDU level, we compared two weighted schemes (self-attentive \cite{vaswani2017attention} and GRU-attentive) with the averaging scheme (see Table \ref{d-level-analysis}). Since these approaches perform similarly, we choose the non-parametric averaging operation for its higher computational efficiency.

\noindent\textbf{Document-Level Encoding}: We also evaluated different document encoder designs: (1) Averaging, (2) Transformer, and (3) Bi-GRU. 
As shown in Table \ref{d-level-analysis}, Transformer and Bi-GRU outperform Averaging, demonstrating the importance of modeling contextual dependencies among EDUs. 
In addition, as shown in Table \ref{before-after-analysis}, adding boundary information to EDU representations after document-level encoding performs better; the relation classification benefits more on utilizing the multi-granularity representation.  

\begin{table}
\linespread{1.1}
\centering
\small
\begin{tabular}{p{3.5cm}cccc}
\hline
\textbf{Model} & \textbf{S} &\textbf{NS} & \textbf{R}  & \textbf{Full}\\
\hline
E-Level Self-Attentive & 87.5 & 75.3 & 63.3 & 62.8\\
E-Level GRU-Attentive & 87.4 & 75.2 & 63.4 & 62.9\\
\hline
D-Level Averaging & 86.9 & 73.5 & 61.1  & 60.6\\
D-Level Transformer & 87.4 & 74.9 & 63.0  &62.5\\
\hline
Proposed Model & 87.9 & 75.4 & 63.3 &62.9\\
\hline
\end{tabular}
\caption{Micro-F1 scores on different EDU-Level (E-Level) and Document-Level (D-Level) encoding methods. The proposed model uses E-Level Averaging and D-Level Bi-GRU.}
\label{d-level-analysis}
\end{table}

\begin{table}
\linespread{1.1}
\centering
\small
\begin{tabular}{lcccc}
\hline
\textbf{Model} & \textbf{S} &\textbf{NS} & \textbf{R}  & \textbf{Full}\\
\hline
Add Boundary Information \\ Before D-Level Encoder
& 86.4 & 73.0 & 59.7 &59.2\\
\hline
Add Boundary Information \\ After D-Level Encoder & 87.9 & 75.4 & 63.3 &62.9\\
\hline
\end{tabular}
\caption{Micro-F1 scores on different layer settings of adding boundary information.}
\label{before-after-analysis}
\end{table}

\subsection{Error Analysis on Top Layer Splitting}
To evaluate the capability of lengthy span splitting, we conducted an ablation experiment of top layer splitting, which is the most challenging decoding step in document-level parsing as discussed in Section \ref{sec:layer_beam_search}. As shown in Table \ref{first-accuracy-analysis}, the adapted top-down sentence-level parser \citep{lin-etal-2019-unified} only reaches 0.263 on the top layer splitting of document-level parsing, while our final model can achieve 0.474. In particular, both layer-wise beam search and boundary information contribute to obtaining more accurate span splitting determination.

\begin{table}[ht]
\linespread{1.1}
\centering
\small
\begin{tabular}{p{3.8cm}ccc}
\hline
\textbf{Model} & \textbf{Accuracy} \\
\hline
Proposed Model & 0.474 \\
- Layer-Wise Beam Search & 0.421 \\
- Boundary Information & 0.289 \\
\citep{lin-etal-2019-unified} & 0.263 \\
\hline
\end{tabular}
\caption{Comparison of top layer splitting accuracy.}
\label{first-accuracy-analysis}
\vspace{-0.2cm}
\end{table}

\section{Conclusion}
We proposed to exploit robust representations of multiple levels of granularity at the syntactic and semantic levels and in turn incorporated such representations in an end-to-end encoder-decoder neural architecture for resourceful discourse processing. Our document-level discourse parser compares favorably with the current state-of-the-art. Experimental results show that our document-based neural discourse parser benefits the most from incorporating boundary information at the EDU level and from modeling global information.

\bibliographystyle{acl_natbib}
\bibliography{emnlp2020}

\end{document}